# Developing and Refining a Multifunctional Facial Recognition System for Older Adults with Cognitive Impairments: A Journey Towards Enhanced Quality of Life

Li He[1], Arizona State University[2]
{lihe5}@asu.edu

## Abstract

In an era where the global population is aging significantly, cognitive impairments among the elderly have become a major health concern. The need for effective assistive technologies is clear, and facial recognition systems are emerging as promising tools to address this issue. This document discusses the development and evaluation of a new Multifunctional Facial Recognition System (MFRS), designed specifically to assist older adults with cognitive impairments. The MFRS leverages face_recognition [1], a powerful open-source library capable of extracting, identifying, and manipulating facial features. Our system integrates the face recognition and retrieval capabilities of face_recognition, along with additional functionalities to capture images and record voice memos. This combination of features notably enhances the system's usability and versatility, making it a more user-friendly and universally applicable tool for end-users. The source code for this project can be accessed at https://github.com/Li-8023/Multi-function-face-recognition.git.

## 1. Introduction

The global demographic landscape is changing dramatically with an increasing percentage of seniors, a significant portion of whom have been diagnosed with Alzheimer's disease. According to an autonomous publication by the U.S. Centers for Disease Control and Defense, it has been learned that an estimated 5.1 million Americans over the age of 65 may currently be suffering from Alzheimer's disease. This number is expected to rise to 13.2 million by 2050 [2]. The prevalence of Alzheimer's disease in the elderly population has led to an urgent need for innovative solutions to address the challenges they face. Among these challenges, impaired facial recognition is a key factor affecting communication, emotional exchange, and safety.

Alzheimer's disease is a neurological condition that worsens over time and is characterized by memory, thinking, and behavior problems. The loss of facial recognition abilities is among the condition's most upsetting side effects since it affects how well people connect and communicate with one another. Alzheimer's patients frequently experience feelings of loneliness and frustration as a result of their inability to recognize familiar people. The difficulties for both dementia patients and their carers are exacerbated by the deterioration in facial recognition skills, which impairs patient interactions and may cause emotional anguish. Additionally, misinterpretations brought on by unfamiliar faces might result in issues like distraction, confusion, and communication challenges, raising safety concerns.

The urgent need for efficient assistive technologies is highlighted by the rise in the world's aging population and the incidence of cognitive impairments like Alzheimer's disease and dementia. These technologies can enhance older individuals' independence and quality of life by offering crucial assistance with daily tasks and memory recall. Facial recognition systems have become one of these technologies' most promising applications. They can support older persons in seamlessly and intuitively recalling the people in their lives.

Over the past few years, facial recognition technology has advanced quickly in the field of computer vision. This is largely because of developments in machine learning and artificial intelligence, which have improved the precision and effectiveness of face recognition from photos and videos. To this progress, the open-source library face_recognition has significantly contributed. It allows researchers and developers all over the world to take advantage of cutting-edge facial recognition technology and is renowned for its simplicity and convenience of use. It uses deep learning to provide robust and excellent facial recognition capabilities, more specifically a Histogram of Oriented Gradients (HOG) based model from dlib. This covers detecting various faces, extracting facial traits from pictures, and even modifying faces in photographs.

However, when it comes to assisting older persons with cognitive problems, the actual application of this technology necessitates a more comprehensive strategy. It involves more than just identifying faces; it also entails offering a user experience that is simple, interesting and catered to the requirements of a particular user group. The system should also have a range of functions to cater to the various demands of elderly people with cognitive impairments. For instance, it entails more than just recognizing a face; it also entails associating that face with pertinent information, such as the person's name or a reminder of any details that might make it easier to recall. As a result, even though this face-recognition library offers a solid foundation for face identification, adding more functionality is necessary for a really successful solution for older persons.

We created the revolutionary Multi-functional Face Recognition System to precisely cater to the demands of older persons with cognitive impairments. MFRS combines the capacity to take images and record voice recordings with face recognition and information retrieval features. The system is more user-friendly and multipurpose because of this feature combination, which takes into account the various demands and capabilities of older persons.

## 2. Related Work

For many years, face recognition has been the focus of in-depth study. Traditional image processing methods like Eigenfaces [3] and Fisherfaces [4], which use linear algebra to extract useful features from images, were widely included in early systems. These techniques are, however, susceptible to changes in lighting, posture, and facial expression. Especially under practical circumstances, their performance is frequently constrained. Recent methods use machine learning, particularly deep learning, to perform facial recognition tasks accurately. These include Google's FaceNet [6] and Facebook's DeepFace [5]. These systems extract high-dimensional information from faces using convolutional neural networks (CNNs), which are then applied to recognition challenges. The face_recognition library has advanced significantly in the area of open source software. The library expands upon the cutting-edge face recognition technology created by dlib using deep learning, offering a very open and adaptable framework for creating face recognition applications [7].

However, research into the application of facial recognition technology as a support system for those with cognitive disabilities is still in its infancy. Its potential has been looked into in a few research. For instance, a robotic companion for adults with cognitive disabilities was developed in a study by Taphs et al. [8] that employs face recognition to distinguish between various people and respond appropriately. However, rather than the facial recognition technology itself, this study concentrated on the human-robot interaction. A facial recognition system was created in a study by Doughty et al. [9] to assist persons with prosopagnosia, a disease marked by a lack of facial identification. The device takes pictures using a wearable camera, processes them to identify faces, and then informs the user audibly about the person they are seeing more of. These systems do not, however, typically feature an integrated strategy to assist older persons with cognitive impairment and instead tend to concentrate on a particular area. To fill this vacuum in the literature, we propose a multifunctional face recognition system in our work.

## 3. Method

In the forthcoming section, we meticulously delineate the constituents of the proposed functional facial recognition system. As depicted in Figure 1, the system we propose is composed of five integral modules: an image capture module, a voice capture module, a facial recognition module, a data storage module, and an information retrieval module.

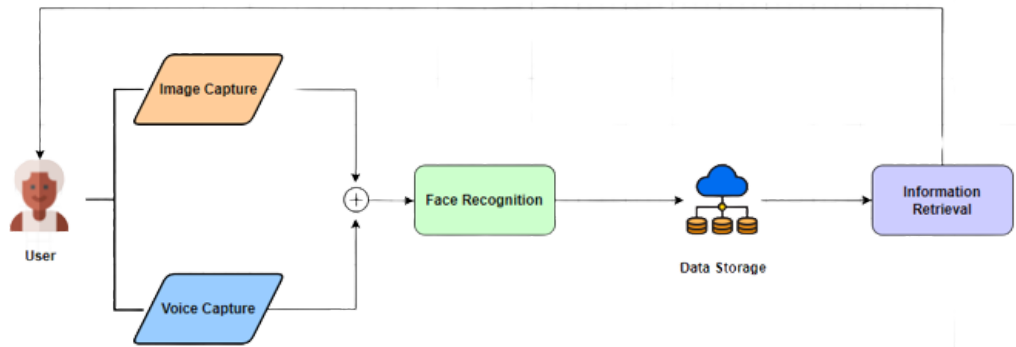

Figure 1. An overview of our MFRS modules

## 3.1. Image Capture Module

The designated module captures images utilizing either an integrated or externally connected camera. It is engineered to operate in conjunction with various types of cameras, thereby ensuring compatibility with a diverse range of devices. The initiation of the camera can be achieved through a simple user command, and the user interface provides a preview window to facilitate appropriate image composition. Upon activation of the camera, the user is enabled to capture images of a novel individual. To guarantee a clear and properly framed facial view within the photograph, the system proffers visual feedback. Should the necessity arise, the user possesses the ability to revisit and scrutinize the captured image.

## 3.2. Voice Capture Module

Users who prefer or find it easier to retain auditory information can use the Voice Memo Recording Module, a special feature of MFRS designed to increase the system's knowledge retrieval capabilities using auditory cues. Users can record any significant details about the interaction as voice memos as soon as they meet new people. For initiating, stopping, and watching recordings, the system offers a straightforward and user-friendly interface. The module is made to function with either the device's built-in microphone or an external microphone that is connected. It can handle a variety of audio qualities and has noise reduction technology to make sure voice memos are audible and comprehensible.

Each voice memo is identified and connected to a particular person in the system database via tags. Based on the setting in which the message was written, the association was made. For instance, if a user instantly records a memo after taking a photo of a brand-new individual, the system links the message to that person. Memos that have been recorded are kept locally on the user's device in an effective audio format that balances file size and quality. The user has the option to play the voice memo when MFRS recognizes someone for whom there is an associated voice memo.

Users with visual impairments may find it simpler to remember people based on their voices rather than their faces, therefore the module is pretty helpful to them. As an alternative to text message displays, it is helpful for users who struggle with dyslexia. The Voice Memo Recording Module is a flexible tool in MFRS because it may also be used to capture any other significant auditory information, such as how to pronounce someone's name. The Voice Memo Recording Module improves the overall efficiency of the MFRS, making it a more complete cognitive assistance for senior citizens by adding an audio layer to information retrieval.

### 3.3. Face Recognition Module

A significant component of MFRS is the facial recognition module. It makes use of the face_recognition library, which is well-known for being precise and effective in identifying faces in photos. This face_recognition library is created using cutting-edge machine learning methods and is assisted by deep learning-based models from dlib. The library uses Python or the command line to recognize and manipulate faces using the world's simplest face recognition library. Strong face detection capabilities are part of the face_recognition library. The face recognition module locates the face in the input image by using a function from the face_locations library. The method returns the face's top, right, bottom, and left coordinates. The face detection function efficiently distinguishes between various people by not only locating the face in the image but also handling several faces in the same image.

Following the faces' detection, the module goes on to extract distinctive features from each face in order to produce an "embedding"—a numerical representation of the face, which is done using a function in the library face_encodings. It returns a list of 128-dimensional face encodings (the encoding of each recognized face). Each face encoding efficiently captures the distinctive qualities that give each face its own identity by representing a facial feature in a condensed manner. The embedded representations of recognized individuals kept in the system database are then compared to the generated face codes. This comparison is done using the compare_faces function. This function accepts a list of candidate encodings as well as a list of known face encodings, and it outputs a list of True/False values indicating which known faces match the candidate face. If a match is discovered, the system correctly detects the person, prompting additional modules to retrieve and present pertinent data about the person. The face recognition module acts as the framework of MFRS, enabling it to aid users in remembering and recognizing people by harnessing the strength of the library face_recognition.

### 3.4. Data Storage Module

The MFRS data storage module is made to manage and keep all of the data that the system produces. It stores and processes the data using the robust open-source object-relational database system PostgreSQL [10]. The data storage module's PostgreSQL database structure enables effective storing and retrieval of many MFRS data types. This contains voice memos, facial embeddings, and associated metadata. To maintain data integrity, tables are built for each type of data with the proper data types and constraints. Each identified person is given a special identification (ID), which enables data linkage between several databases of connected information. The ID makes sure that each person's information is efficiently obtained and updated. The data store module updates the PostgreSQL database whenever a new face is identified. The module saves the face embedding and any user-provided metadata for fresh faces, like the person's name, relationships, and other specifics. An interface for data administration operations, such as modifying already-existing data entries, deleting data, and adding data, is provided by the data storage module. Make use of PostgreSQL's robust transaction management and backup features to guarantee the integrity and security of your data.

### 3.5. Information Retrieval Module

When an individual is recognized by the system, the information retrieval module locates and displays pertinent information. It has intimate relationships with both the PostgreSQL database and the data storage module. The Information Retrieval module searches the PostgreSQL database for pertinent data when an individual is recognized. This comprises the person's metadata, any related voice memos, and perhaps other pertinent information. SQL queries are performance-optimized to guarantee quick and effective data retrieval. The robust querying features and indexes of PostgreSQL are employed to speed up data access. The information retrieval module formats the data for presentation after data retrieval. This could entail reading text on the screen or playing voice recordings, depending on the system's layout and user settings.

## 4. Future work

Numerous prospective avenues for future development exist, which could enhance the functionality and usability of the MFRS. Despite the fact that its present implementation provides a robust foundation for facilitating social interactions among older individuals with cognitive limitations, there is always room for further improvements and refinements.

### 4.1 System Assessment

Future work must start with a thorough assessment of the system. To confirm the effectiveness, precision, and usability of the MFRS, a comprehensive assessment is required. Both quantitative and qualitative techniques, such as system testing, user testing, and feedback gathering, will probably be used in this phase.

Thorough system testing is the first step in the evaluation of a system. This can be categorized into two main groups:

1. **Technical assessment:** Testing the system to gauge its technical performance is known as a technical assessment. This includes evaluating the effectiveness of the facial recognition technology and the database operations.
    a. **Evaluation of Face Recognition:** To assess the accuracy of recognition, we will utilize a benchmark dataset (such as the Labeled Faces in the Wild [11] or YouTube Faces [12] datasets). We can compare the outcomes with other cutting-edge facial recognition methods thanks to the usage of well-established datasets.
    b. **Evaluation of database operations:** Stress tests can be run on database operations to see how well the system operates under high load. Calculate how long it takes to insert, retrieve, and update data in the database.

2. **User Interface Evaluation:** Assess the usefulness of the user interface by assessing the system's responsiveness, the design's intuitiveness, and the ease of navigation. Prior to undertaking user testing, this can be completed internally.

## 4.2 User Research

User testing should be done as soon as the system's technical performance has been confirmed. This entails interacting with the system and soliciting feedback from a group of target users regarding their experience.

1. **Usability study:** With a group of older persons (the MFRS target population), conduct a usability study. Ask for input on the user interface and overall experience, observe how they interact with the system, and note any problems they run across.

2. **Longitudinal Study:** If feasible, a longitudinal study can provide a deeper understanding of the system's performance over time. This involves user interaction with the system over an extended period and can provide valuable insights into the system's durability and usefulness under real-life conditions.

## 4.3. Collecting and Analyzing Feedback

After collecting feedback, it should be systematically analyzed. For quantitative data from questionnaires, statistical analysis can be used to reveal patterns and significant findings. For qualitative data from interviews and open-ended survey questions, thematic analysis can be applied to identify recurring themes and patterns. By considering feedback from all these sources, we can gain a comprehensive understanding of the system's strengths and weaknesses from the user's perspective. This information is invaluable for guiding future improvements to the system.

### 4.4. Incorporating additional modalities

While the MFRS currently uses facial recognition for identity verification, integrating other modalities can enhance the system's accuracy and robustness. Adding voice recognition capabilities to the system can be helpful in situations where faces are unclear or lighting is insufficient. Voice recognition can serve as an independent method of identification or can be combined with facial recognition to improve accuracy.

Future work in these areas will enable MFRS to develop further, better serve its target consumers, and broaden the scope of its potential applications.

## 5. Conclusion

The process of creating an effective Multimodal Facial Recognition System for older adults, especially those with cognitive impairments, is a challenging yet rewarding endeavor. The system, capable of identifying familiar individuals and providing contextual information, can help enhance the quality of life for the target population.

However, it's important to remember that this is just the beginning of the journey. The system's full potential can only be realized through a continuous cycle of testing, evaluation, and refinement. This process will involve rigorous technical performance testing of the system, usability testing with the target users, and the collection and analysis of feedback to identify areas for improvement.

Looking forward, there are exciting opportunities to expand the system's capabilities. Integrating additional modalities like voice recognition can enhance the system's robustness and accuracy in identifying individuals. Therefore, while we've made significant progress in developing the MFRS, the journey of refinement and enhancement continues. We are committed to continually improving the system under the guidance of user feedback and technological advancements to maximize benefits for the target population.